%% file: main.tex
\pdfoutput=1

\documentclass[11pt]{article}

\usepackage{misc/acl}

\usepackage[utf8]{inputenc} 
\usepackage[T1]{fontenc}    
\usepackage{times}          
\usepackage{inconsolata}    
\usepackage{pifont}

\usepackage{microtype}      
\usepackage{url}            
\usepackage{hyperref}       

\usepackage{rotating}
\usepackage{makecell}
\usepackage{booktabs}       
\usepackage{array}          
\usepackage{multirow}       
\usepackage{adjustbox}      
\usepackage{colortbl}       
\definecolor{Gray}{gray}{0.98}
\newcolumntype{g}{>{\columncolor{Gray}}c}
\usepackage{tabularx}

\usepackage{fix-cm}

\usepackage[dvipsnames,x11names]{xcolor}
\usepackage{todonotes}

\usepackage{amsmath}        
\usepackage{amssymb}        
\usepackage{amsfonts}       
\usepackage{mathtools}      
\usepackage{amsthm}         
\usepackage{bm}
\usepackage{esvect}

\usepackage{cleveref}

\usepackage[inline, shortlabels]{enumitem} 

\usepackage{graphicx}       
\usepackage{subfigure}      
\usepackage{xspace}         

\usepackage[algoruled,ruled,vlined,noend]{algorithm2e} 

\usepackage{soul}           
\usepackage{changepage}     
\input{macros.tex}
\graphicspath{{img/}}

\definecolor{c1}{HTML}{E85642}

\hypersetup{
}

\title{Unsupervised Word-level Quality Estimation for Machine Translation \\ Through the Lens of Annotators (Dis)agreement}

\author{
  Gabriele Sarti$^1$ ~\;~ Vilém Zouhar$^2$ ~\;~ \textbf{Malvina Nissim}$^1$ ~\;~ \textbf{Arianna Bisazza}$^1$\vspace{3mm}\\
  $^1$CLCG, University of Groningen ~\;~ $^2$ETH Zurich \vspace{3mm}\\
  \small{\texttt{\{\href{mailto:g.sarti@rug.nl}{\color{black} g.sarti}, \href{mailto:a.bisazza@rug.nl}{\color{black} a.bisazza}\}@rug.nl}}
}

\begin{document}
\maketitle

\begin{abstract}

Word-level quality estimation (WQE) aims to automatically identify fine-grained error spans in machine-translated outputs and has found many uses, including assisting translators during post-editing. Modern WQE techniques are often expensive, involving prompting of large language models or ad-hoc training on large amounts of human-labeled data. In this work, we investigate efficient alternatives exploiting recent advances in language model interpretability and uncertainty quantification to identify translation errors from the inner workings of translation models. In our evaluation spanning 14 metrics across 12 translation directions, we quantify the impact of human label variation on metric performance by using multiple sets of human labels. Our results highlight the untapped potential of unsupervised metrics, the shortcomings of supervised methods when faced with label uncertainty, and the brittleness of single-annotator evaluation practices.

\end{abstract}

\blankfootnote{$^\dagger$Materials: \href{https://github.com/gsarti/labl/tree/main/examples/unsup_wqe}{\texttt{gsarti/labl/examples/unsup\_wqe}.}}

\section{Introduction}

Word-level error spans are widely used in machine translation (MT) evaluation to obtain robust and fine-grained estimates of translation quality \citep{lommel-etal-2014-using,freitag-etal-2021-experts, freitag-etal-2021-results, kocmi-etal-2024-error}.
Due to the cost of manual annotation, word-level quality estimation (WQE) was proposed for assisting in annotating error spans over MT outputs \citep{zouhar-etal-2025-ai}.
%
%
Modern WQE approaches generally rely on costly inference with large language models (LLMs) or ad-hoc training with large amounts of human-annotated texts \citep{fernandes-etal-2023-devil,kocmi-federmann-2023-large,guerreiro-etal-2024-xcomet}, making them impractical for less resourced settings \citep{zouhar-etal-2024-fine}.
To improve the efficiency of MT quality assessment, several works explored the use of signals derived from the internals of neural MT systems \citep{fomicheva-etal-2020-unsupervised,fomicheva-etal-2021-eval4nlp,leiter-etal-2024-towards}, for identifying problems in MT outputs, such as hallucinations \citep{guerreiro-etal-2023-optimal,guerreiro-etal-2023-looking,dale-etal-2023-detecting,dale-etal-2023-halomi,himmi-etal-2024-enhanced}.
However, previous works focus on sentence-level metrics for overall translation quality, and do not evaluate performance on multiple label sets due to high annotation costs \citep{fomicheva-etal-2022-mlqe,zerva-etal-2024-findings}.\footnote{Other relevant works are discussed in~\Cref{app:related-work}}

\begin{figure}
    \centering
    \includegraphics[width=\linewidth]{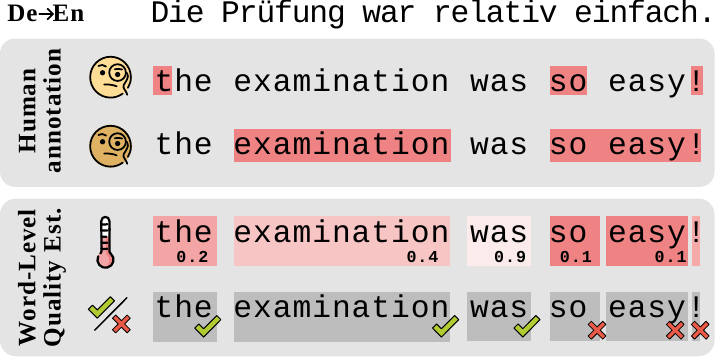}
    \caption{Example of German$\rightarrow$English translation with two sets of human word-level error span annotations and two examples of continuous and binary WQE metrics.}
    \label{fig:intro_fig}
    \vspace{-10pt}
\end{figure}

In this work, we conduct a more comprehensive evaluation spanning 10 unsupervised metrics derived from models' inner representations and predictive distributions to identify translation errors at the word level. We test three open-source multilingual MT models and LLMs of different sizes across 12 translation directions, including typologically diverse languages and challenging textual domains.
Importantly, we focus on texts with \textit{multiple} human annotations to measure the impact of individual annotator preferences on metric performance, setting a ``human-level'' baseline for the WQE task.

We address the following research questions:
\textbf{i)}~How accurate are unsupervised WQE metrics in detecting MT errors compared to trained metrics and human annotators?
\textbf{ii)}~Are popular supervised WQE metrics well-calibrated?
\textbf{iii)}~Are the relative performances of WQE metrics affected by the variability in human error annotations?
We conclude with recommendations for improving the evaluation and usage of future WQE systems.

\begin{table}
    \footnotesize
    \centering
    \begin{tabular}{m{1.6cm}m{5em}m{4em}m{4em}}
    \toprule[1.5pt]
     &  \b{DivEMT} & \b{WMT24} & \b{QE4PE} \\
    \midrule
    \b{Languages} & \textsc{en}$\rightarrow$\textsc{ar,it, nl,tr,uk,vi} & \textsc{en}$\rightarrow$\textsc{ja,zh,
    hi,cs,ru} \CsUk & \textsc{en}$\rightarrow$\textsc{it,nl} \\
    \midrule
    \b{Errors type} & Post-edit & Annotation & Post-edit \\
    \midrule
    \b{Label sets} & 1 & 1 & 6 \\
    \midrule
    \b{Domains} & Wiki & Multiple & Social, Biomed \\
    \midrule
    \b{MT Model} & mBART-50 & Aya23 & NLLB \\
    \midrule
    \b{\# Segments} & 2580 & 5124 & 3888 \\
    \bottomrule[1.5pt]
    \end{tabular}
    \caption{
    Summary of tested datasets.
    Error spans are obtained from explicit error annotations or post-edited spans.
    Additional details are available in~\Cref{app:data}.}
    \label{tab:datasets-summary}
    \vspace{-12pt}
\end{table}

\section{Data}
We use datasets containing error annotations or post-edits on the outputs of open-source models to extract unsupervised WQE metrics on real model outputs, avoiding possible confounders.
We select the following datasets, summarized in \Cref{tab:datasets-summary}:

\vspace{5pt}\noindent
\textbf{DivEMT} \citep{sarti-etal-2022-divemt} contains a single set of post-edits over translations produced by mBART-50 \citep{tang-etal-2021-multilingual} for a subset of Wiki texts from the FLORES dataset \citep{goyal-etal-2022-flores} spanning six typologically diverse target languages (\textsc{en}$\rightarrow$\textsc{ar,it,nl,tr,uk,vi}). We use it for a cross-lingual comparisons over a fixed set of examples.

\vspace{5pt}\noindent
\textbf{WMT24} \citep{kocmi-etal-2024-findings} contains error spans on the outputs of the Aya23-35B LLM \citep{aya23} produced for the WMT24 General Translation Shared Task spanning multiple domains across six directions (\textsc{en}$\rightarrow$\textsc{ja,zh,hi,cs,ru} and \CsUk). It was selected to extend our evaluation to a state-of-the-art LLM, given the popularity of such systems in MT~\citep{kocmi-etal-2023-findings}.

\vspace{5pt}\noindent
\textbf{QE4PE} \citep{sarti-etal-2025-qe4pe} contains multiple human professional post-edits over translations produced by the NLLB 3.3B model \citep{nllb} for \EnIt and \EnNl on challenging textual domains (social posts and biomedical abstracts). This dataset is used to conduct our evaluation across multiple annotation sets.


\section{Evaluated Metrics}

The following metrics were evaluated using the Inseq library \citep{sarti-etal-2023-inseq,sarti-etal-2024-democratizing}.

\paragraph{Predictive Distribution Metrics.}
We use the \textbf{Surprisal} of the predicted token $t^{*}$, as negative log-probablity $-\log p(t^{*}_i|t_{<i})$, and the \textbf{Entropy} $H$ of the output distribution $P_N$ over vocabulary $V$, $-\sum_{i=1}^{|V|} p(t_i|t_{<i}) \log_2 p(t_i|t_{<i})$, as simple metrics to quantify pointwise and full prediction uncertainty \citep{fomicheva-etal-2020-unsupervised}. For surprisal, we also compute its expectation (\textbf{MCD\textsubscript{\textsc{avg}}}) and variance (\textbf{MCD\textsubscript{\textsc{var}}}) with $n=10$ steps of Monte Carlo Dropout (MCD, \citealp{mcdropout}) to obtain a robust estimate and a measure of epistemic uncertainty in predictions, respectively.\footnote{Epistemic uncertainty reflects models' lack of knowledge rather than data ambiguity. MCD is tested only on encoder-decoder models since Aya layers do not include dropout.}

\paragraph{Vocabulary Projections.}
We use the LogitLens (LL, \citealp{logitlens}) to extract probability distributions $P_0, \dots, P_{N-1}$ over $V$ from intermediate activations at every layer $l_0, \dots, l_{N-1}$ of the decoder.
We use the surprisal for the final prediction at every layer (\textbf{LL-Surprisal}) to assess the presence of layers with high sensitivity to wrong predictions.
Then, we compute the KL divergence between every layer distribution and the final distribution $P_N$, e.g. $\text{KL}(P_{N-1}\|P_N)$, to highlight trends in the shift in predictive probability produced by the application of remaining layers (\textbf{LL KL-Div}).
Finally, we adapt the approach of \citet{prediction-depth} and use the number of the first layer for which the final prediction corresponds to the top logit as a metric of model confidence, $l \;\text{s.t.}\;\arg \max P_l = t^{*}$ and $\arg \max P_i \neq t^{*} \;\forall i<l$ (\textbf{LL Pred. Depth}).

\paragraph{Context mixing.}
We use the entropy of the distribution of attention weights\footnote{For encoder-decoder model, self-attention and cross-attention weights are concatenated and renormalized.} over previous context as a simple measure of information locality during inference \citep{ferrando-etal-2022-measuring,mohebbi-etal-2023-quantifying}.
Following \citet{fomicheva-etal-2020-bergamot}, we experiment with using the mean and the maximum entropy across all attention heads of all layers as separate metrics (\textbf{Attn. Entropy\textsubscript{\textsc{var/max}}}).
Finally, we evaluate the Between Layer OOD method by \citet{jelenic2024blood}, employing gradients to estimate layer transformation smoothness for OOD detection (\textbf{\textsc{Blood}}).

\paragraph{Supervised baselines.}
We also test the state-of-the-art supervised WQE model \textsc{xcomet} \citep{guerreiro-etal-2024-xcomet} in its XL (3.5B) and XXL (10.7B) sizes, using them as binary metrics.
Contrary to the continuous metrics from the previous section, binary labels from \textsc{xcomet} cannot be easily calibrated to match subjective annotation propensity. Hence, we propose to adapt the \textsc{xcomet} metric to use the sum of probability for all error types as a token-level continuous confidence metric, $s(t^{*}) = p(\text{\textsc{minor}}) + p(\text{\textsc{major}}) + p(\text{\textsc{critical}})$, which we dub \textbf{\textsc{xcomet}\textsubscript{\textsc{conf}}}.

\paragraph{Human Editors.}
For QE4PE, we report the min/mean/max agreement between each annotator's edited spans and those of the other five editors as a less subjective ``human-level'' quality measure. 

\setlength{\tabcolsep}{4.5pt}
\begin{table}
    \small
    \centering

\begin{tabular}{l@{\hspace{1mm}}lcccccc}
    \toprule[1.5pt]
    & \multirow{2}{*}{\b{Method}} & \multicolumn{2}{c}{\b{DivEMT}} & \multicolumn{2}{c}{\b{WMT24}} & \multicolumn{2}{c}{\b{QE4PE}} \\

    \cmidrule(lr){3-4}
    \cmidrule(lr){5-6}
    \cmidrule(lr){7-8}
                                    & & \b{AP}  & \b{F1$^{*}$}  & \b{AP}  & \b{F1$^{*}$}  & \b{AP}  & \b{F1$^{*}$}  \\
    \midrule
    & Random                       &    \cellcolor{PaleGreen3!0}{.34}  & \cellcolor{DeepSkyBlue3!38}{.50}     & \cellcolor{PaleGreen3!0}{.05}   & \cellcolor{DeepSkyBlue3!0}{.09}   & \cellcolor{PaleGreen3!3}{.17}     & \cellcolor{DeepSkyBlue3!0}{.27}   \\
    \midrule
    \multirow{10}{*}{\rotatebox[origin=c]{90}{\textsc{unsupervised}}} & Surprisal                    &    \cellcolor{PaleGreen3!24}{.43}  & \cellcolor{DeepSkyBlue3!51}{.53}     & \cellcolor{PaleGreen3!16}{.08}   & \cellcolor{DeepSkyBlue3!16}{.13}   & \cellcolor{PaleGreen3!24}{.23}     & \cellcolor{DeepSkyBlue3!15}{.32}     \\
    & Out. Entropy                 &    \cellcolor{PaleGreen3!32}{.46}  & \cellcolor{DeepSkyBlue3!42}{.51}    &  \u{\cellcolor{PaleGreen3!27}{.10}}   & \u{\cellcolor{DeepSkyBlue3!28}{.16}}   & \cellcolor{PaleGreen3!24}{.23}     & \cellcolor{DeepSkyBlue3!11}{.31}     \\
    & Surprisal \textsc{mcd}\avg   &    \cellcolor{PaleGreen3!24}{.43}  & \cellcolor{DeepSkyBlue3!51}{.53}     & \cellcolor{white}{-}     & \cellcolor{white}{-}     & \cellcolor{PaleGreen3!28}{.24}     & \cellcolor{DeepSkyBlue3!18}{.33}     \\
    & Surprisal \textsc{mcd}\var   & \u{\cellcolor{PaleGreen3!35}{.47}} & \u{\cellcolor{DeepSkyBlue3!55}{.54}} & \cellcolor{white}{-}     & \cellcolor{white}{-}     & \u{\cellcolor{PaleGreen3!35}{.26}} & \u{\cellcolor{DeepSkyBlue3!21}{.34}} \\
    & LL Surprisal\best            &    \cellcolor{PaleGreen3!21}{.42}  & \cellcolor{DeepSkyBlue3!51}{.53}     & \cellcolor{PaleGreen3!21}{.09}   & \cellcolor{DeepSkyBlue3!24}{.15}   & \cellcolor{PaleGreen3!24}{.23}     & \cellcolor{DeepSkyBlue3!15}{.32}     \\
    & LL KL-Div\best               &    \cellcolor{PaleGreen3!24}{.43}  & \cellcolor{DeepSkyBlue3!42}{.51}     & \cellcolor{PaleGreen3!10}{.07}   & \cellcolor{DeepSkyBlue3!12}{.12}   & \cellcolor{PaleGreen3!14}{.20}     & \cellcolor{DeepSkyBlue3!5}{.29}     \\
    & LL Pred. Depth               &    \cellcolor{PaleGreen3!13}{.39}  & \cellcolor{DeepSkyBlue3!42}{.51}     & \cellcolor{PaleGreen3!5}{.06}   & \cellcolor{DeepSkyBlue3!12}{.12}   & \cellcolor{PaleGreen3!14}{.20}     & \cellcolor{DeepSkyBlue3!5}{.29}     \\
    & Att. Entropy\avg             &    \cellcolor{PaleGreen3!8}{.37}  & \cellcolor{DeepSkyBlue3!38}{.50}     & \cellcolor{PaleGreen3!0}{.05}   & \cellcolor{DeepSkyBlue3!0}{.09}   & \cellcolor{PaleGreen3!7}{.18}     & \cellcolor{DeepSkyBlue3!3}{.28}     \\
    & Att. Entropy\maxim           &    \cellcolor{PaleGreen3!0}{.34}  & \cellcolor{DeepSkyBlue3!38}{.50}     & \cellcolor{PaleGreen3!0}{.05}   & \cellcolor{DeepSkyBlue3!0}{.09}   & \cellcolor{PaleGreen3!0}{.16}     & \cellcolor{DeepSkyBlue3!3}{.28}     \\
    & \textsc{Blood}\best          &    \cellcolor{PaleGreen3!0}{.34}  & \cellcolor{DeepSkyBlue3!38}{.50}     & \cellcolor{white}{-}     & \cellcolor{white}{-}     & \cellcolor{PaleGreen3!3}{.17}     & \cellcolor{DeepSkyBlue3!3}{.28}     \\
    \midrule
    \multirow{4}{*}{\rotatebox[origin=c]{90}{\textsc{super.}}} & \textsc{xcomet-xl}           &    \cellcolor{PaleGreen3!21}{.42}  & \cellcolor{DeepSkyBlue3!17}{.45}     & \cellcolor{PaleGreen3!21}{.09}   & \cellcolor{DeepSkyBlue3!40}{.19}   & \cellcolor{PaleGreen3!24}{.23}     & \cellcolor{DeepSkyBlue3!21}{.34}     \\
    & \textsc{xcomet-xl}\confw     &    \cellcolor{PaleGreen3!54}{.54}  & \b{\cellcolor{DeepSkyBlue3!60}{.55}} & \cellcolor{PaleGreen3!54}{.15}   & \cellcolor{DeepSkyBlue3!56}{.23}   & \cellcolor{PaleGreen3!56}{.32}     & \b{\cellcolor{DeepSkyBlue3!30}{.37}} \\
    & \textsc{xcomet-xxl}          &    \cellcolor{PaleGreen3!24}{.43}  & \cellcolor{DeepSkyBlue3!0}{.41}     & \cellcolor{PaleGreen3!21}{.09}   & \cellcolor{DeepSkyBlue3!44}{.20}   & \cellcolor{PaleGreen3!21}{.22}     & \cellcolor{DeepSkyBlue3!11}{.31}     \\
    & \textsc{xcomet-xxl}\confw    & \b{\cellcolor{PaleGreen3!60}{.56}} & \b{\cellcolor{DeepSkyBlue3!60}{.55}} & \b{\cellcolor{PaleGreen3!60}{.16}}   & \b{\cellcolor{DeepSkyBlue3!60}{.24}}   & \b{\cellcolor{PaleGreen3!60}{.33}} & \b{\cellcolor{DeepSkyBlue3!30}{.37}} \\
    \midrule
    \multirow{3}{*}{\rotatebox[origin=c]{90}{\textsc{hum.}}} & Hum. Editors\minim          & \cellcolor{white}{-}       & \cellcolor{white}{-}       & \cellcolor{white}{-}     & \cellcolor{white}{-}     & \cellcolor{PaleGreen3!28}{.24}     & \cellcolor{DeepSkyBlue3!21}{.34}     \\
    & Hum. Editors\avg            & \cellcolor{white}{-}       & \cellcolor{white}{-}       & \cellcolor{white}{-}     & \cellcolor{white}{-}     & \cellcolor{PaleGreen3!42}{.28}     & \cellcolor{DeepSkyBlue3!42}{.41}     \\
    & Hum. Editors\maxim          & \cellcolor{white}{-}       & \cellcolor{white}{-}       & \cellcolor{white}{-}     & \cellcolor{white}{-}     & \cellcolor{PaleGreen3!56}{.32}     & \cellcolor{DeepSkyBlue3!60}{.47}     \\
    \bottomrule[1.5pt]
\end{tabular}
    
    \caption{Average Precision (\hlc[PaleGreen3!60]{AP}) and Optimal F1 (\hlc[DeepSkyBlue3!50]{F1$^{*}$}) for metrics across tested datasets. Results are averaged across all languages and annotators, with \u{best unsupervised} and \b{overall best} results highlighted.}
    \label{tab:datasets-perf}
    \vspace{-12pt}
\end{table}

\section{Experiments}

\paragraph{How Accurate are Unsupervised WQE Metrics?} \Cref{tab:datasets-perf} reports the average metrics performance across all translation directions across tested datasets.\footnote{Full breakdown available in the Appendix (\Cref{tab:qe4pe-ita-all-results,tab:qe4pe-nld-all-results,tab:divemt-all-results,tab:wmt24esa-all-results}).} We report Average Precision (AP) as a general measure of metric quality across the full score range, and we estimate calibrated metric performance as the best F1 score (F1$^{*}$) across all thresholds for binarizing continuous metric scores into pos./neg. labels matching human annotation.\footnote{Random baseline AP values match the proportion of tokens marked as errors, which can vary greatly.} Our results show that, despite high variability in error span prevalence across different models, languages and annotators, metric rankings remain generally consistent, suggesting that \finding{variability in does not affect the general ability of best metrics to discern underlying phenomena} resulting in translation errors.
Among unsupervised metrics, we find those based on the output distribution to be most effective at identifying error spans, in line with previous segment-level QE results \citep{fomicheva-etal-2020-unsupervised}. Notably, the Surprisal MCD\textsubscript{\textsc{var}} shows strong performances in line with the default \textsc{xcomet} models. For the multi-label QE4PE dataset, we find that the best supervised metrics score on par with the average human annotator consensus (Hum. Editors\textsubscript{\textsc{avg}}), while unsupervised metrics generally obtain lower performances.

\begin{figure}
    \centering
    \includegraphics[width=0.69\linewidth, trim=2mm 0mm 0mm 0mm, clip]{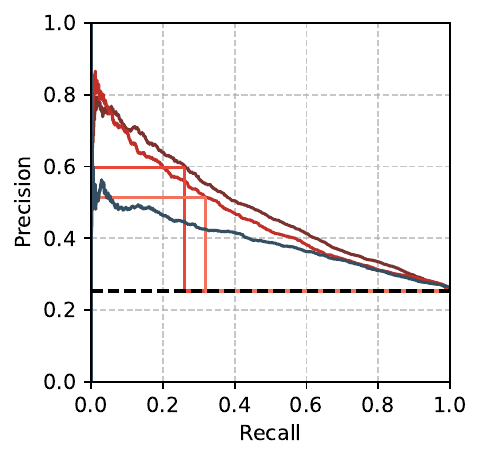}
    \hspace{-3.5mm}
    \includegraphics[width=0.31\linewidth, trim=2mm 8mm 4mm 2mm, clip]{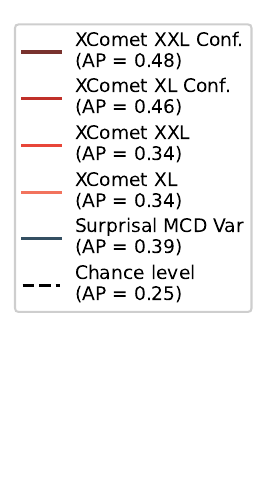}
    
    \caption{Precision-Recall tradeoff for binary and confidence-weighted \textsc{xcomet} variants and the Surprisal \textsc{mcd}\textsubscript{\textsc{var}} metric for DivEMT \EnIt.}
    \label{fig:pr-curves-main}
    \vspace{-12pt}
\end{figure}

\paragraph{Confidence Weighting Enables \textsc{xcomet} Calibration.}
From~\Cref{tab:datasets-perf} results, default \textsc{xcomet} metrics underperform compared to the best unsupervised techniques, a surprising result given their ad-hoc tuning. On the contrary, our \textsc{xcomet}\textsubscript{\textsc{conf}} method consistently reaches better results across all tested sets. \Cref{fig:pr-curves-main} shows the precision-recall tradeoff for these metrics on the \EnIt subset of the \textsc{DivEMT} dataset.\footnote{Results for all datasets in the Appendix (\Cref{fig:qe4pe-ita-pr-curves,fig:qe4pe-nld-pr-curves,fig:divemt-pr-curves,fig:wmt24esa-pr-curves}).} In their default form commonly used for evaluation via the \href{https://pypi.org/project/unbabel-comet/}{\color{black} \texttt{unbabel-comet}} library, \textsc{xcomet} metrics consistently outperform Surprisal MCD\textsubscript{\textsc{var}} in terms of precision (51-60\%, compared to 34\% optimal precision for MCD\textsubscript{\textsc{var}}), but identify only 32-26\% of tokens annotated as errors, resulting in lower AP.
The low recall of these metrics might be problematic in WQE applications where omitting an error might result in oversights from human post-editors trusting the comprehensiveness of WQE predictions. On the contrary, the confidence-weighted \textsc{xcomet}\textsubscript{\textsc{conf}} show strong performances across the whole recall range, resulting in consistent improvements in both F1$^{*}$ and AP~\Cref{tab:datasets-perf}. Concretely, these results confirm that default \textsc{xcomet} performance does not reflect the full capacity of the metric, and \finding{operating with granular confidence scores can be beneficial when calibration is possible}.

\paragraph{Metrics Performance for Multiple Annotations.}
While our evaluation so far employed human error span annotations as binary labels, we set out to assess how more granular labeling schemes impact metrics' performance. Given $L$ sets of binary labels (up to 6 per language for QE4PE), we assign a score $s \in \{1,\dots,L\}\;$ to every MT token using the number of annotators that marked it as an error, resulting in edit counts reflecting human agreement rate.\footnote{An example is available in the Appendix (\Cref{tab:example-qe4pe-ita}).}
\Cref{fig:annotators-perf} presents the correlation of various metrics when the number of annotators available is increased, with median values and confidence bounds are obtained from edit counts across all combinations of $L$ label sets.\footnote{$x$=1 corresponds to binary labels from previous sections.} The increasing trend for correlations across all reported metrics indicates that these methods reflect well the \textit{aleatoric uncertainty} in error span labels, i.e. the disagreement between various annotators. In particular, the Surprisal MCD\textsubscript{\textsc{var}} metric sees a steeper correlation increase than other well-performing metrics, surpassing default \textsc{xcomet} supervised approaches for higher correlation bins. This suggests the epistemic uncertainty derived from noisy model predictions might be a promising way to anticipate the aleatoric uncertainty across human annotators for WQE. We observe that 95\% confidence intervals for high-scoring metrics are largely overlapping when a single set of labels is used, indicating that \finding{rankings of metric performance are subject to change depending on subjective choices of the annotator}. While this poses a problem when attempting a robust evaluation of WQE metrics, we remark that including multiple annotations largely mitigates this issue. As a result, we recommend to explicitly account for human label variation by including multiple error annotations in future WQE evaluations to ensure generalizable findings.

\begin{figure}
    \centering
    \includegraphics[width=\linewidth]{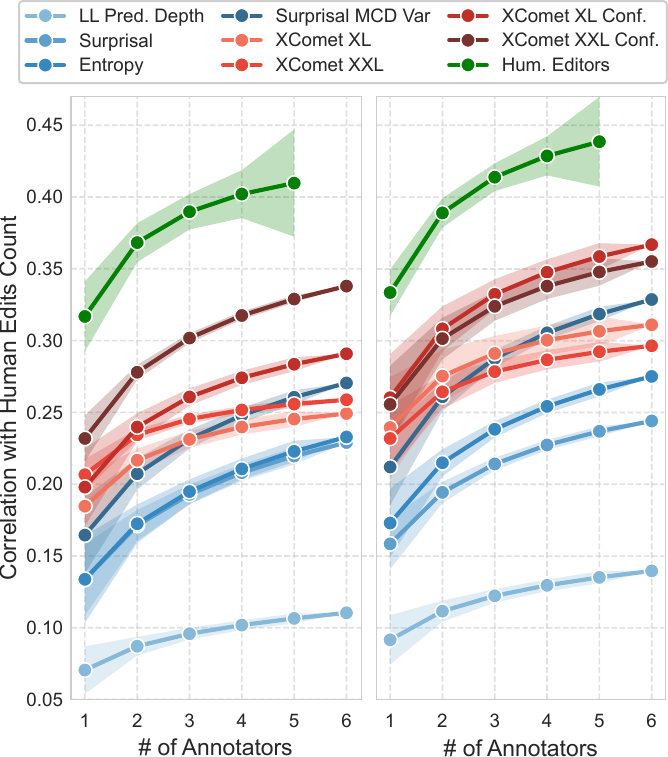}
    \caption{Spearman correlation between WQE metric scores and human edit counts across multiple annotation sets for QE4PE \EnIt (left) and \EnNl (right).}
    \label{fig:annotators-perf}
    \vspace{-12pt}
\end{figure}

\section{Conclusion}

We conducted a comprehensive evaluation of supervised and unsupervised WQE metrics across multiple languages and annotation sets.
Our results show that \textbf{i)} While unsupervised metrics generally lag behind state-of-the-art supervised systems, some uncertainty quantification methods based on the predictive distribution show promising correlation with human label variation;
\textbf{ii)} Popular supervised WQE metrics have generally low levels of recall, and can benefit from confidence weighting to when calibration is possible;
and \textbf{iii)} Individual annotator preferences are key confounders in WQE evaluations and can be mitigated by making use of multiple annotation sets.
We offer the following practical recommendations for evaluating WQE systems:
\begin{itemize}[left=0mm,noitemsep,topsep=0mm]
    \item Use agreement between multiple human annotations to control the effect of subjective preferences and rank WQE metrics robustly.
    \item Employ an in-distribution calibration set of error spans before testing to ensure fair metric comparisons, and favor evaluations accounting for precision-recall tradeoffs to ensure their usability across various confidence levels.
    \item Previous work showed the effectiveness of visualization reflecting prediction confidence~\citep{vasconcelos-etal-2025-generation}, such as highlights for various error severity levels~\citep{sarti-etal-2025-qe4pe}. Consider using continuous WQE metrics in real-world applications such as WQE-augmented post-editing to convey fine-grained confidence variations.
\end{itemize}


\section*{Limitations}

Our findings are accompanied by a number of limitations. Firstly, our choice of tested datasets was limited by the availability of annotated outputs generated by open-source MT models. While several other datasets matching these criteria exist~\citep{fomicheva-etal-2022-mlqe,yang-etal-2023-rethinking,dale-etal-2023-halomi}, we restricted our assessment to a sufficient subset to ensure diversity across languages and tested models to support our findings. To facilitate comparison with other datasets, our evaluation for WMT24 treats available error spans as binary labels and does not directly account for error severity in human-annotated spans. Our choice of unsupervised metrics was largely driven by previous work on uncertainty quantification in MT, and ease of implementation for popular methods in mechanistic interpretability literature~\citep{ferrando2024primerinnerworkingstransformerbased}. However, our choices in the latter category were limited since most methods are nowadays developed and tested specifically for decoder-only transformer models. Finally, despite their strong performance, we found unsupervised methods based on MCD to require substantial computational resources, and as such we could not evaluate them on Aya23 35B. While our main focus was to establish baseline performances across various popular methods, future work should leverage the latest insights from more advanced techniques requiring, for example, the tuning of vocabulary projections~\citep{belrose2023elicitinglatentpredictionstransformers,yom-din-etal-2024-jump} or the identification of ``confidence neurons'' modulating predictive entropy~\citep{confidence-neurons}.

\bibliography{anthology.fixed,custom}
\bibliographystyle{misc/acl_natbib}

\clearpage
\appendix

\section{Additional Background}
\label{app:related-work}

In this section, we provide additional background information regarding the topics of our work.

\paragraph{Unsupervised Quality Estimation for Machine Translation.}

The use of unsupervised signals from MT models for the task of MT quality estimation was introduced by \citet{fomicheva-etal-2020-unsupervised}. Their evaluation showed that high-performing unsupervised methods could rival state-of-the-art supervised QE models in predicting translation quality at the sentence level. Since then, several evaluation campaigns assessed the quality of QE methods~\citep{specia-etal-2021-findings,zerva-etal-2022-findings,blain-etal-2023-findings,zerva-etal-2024-findings}, including a shared task dedicated to explainable QE metrics \citep{fomicheva-etal-2021-eval4nlp}. However, such evaluations have typically focused on segment-level evaluation quality, with word-level error spans being generally obtained by attributing the predictions of supervised segment-level metrics~\citep{rubino-etal-2021-error,rei-etal-2023-inside}. By contrast, recent work on LLMs evaluates various metrics to detect errors from the generator model, without additional systems involved, both at the sentence \citep{fadeeva-etal-2023-lm} and at the token level \citep{fadeeva-etal-2024-fact}. Our work follows the latter approach by testing unsupervised metrics extracted from an MT model during generation, akin to out-of-distribution detection in signal processing research \citep{ood-detection}.

\paragraph{Actionable Insights from Interpretability.} Advances in interpretability research have elucidated multiple mechanisms underlying decision-making, knowledge representation, and biases in LMs \citep{ferrando2024primerinnerworkingstransformerbased}. However, a better understanding of model's inner workings often did not translate to tangible gains in model design and other practical applications, which remain rarely explored~\citep{mosbach-etal-2024-insights}. Some examples in this direction include using targeted machine unlearning methods for safety-critical scenarios~\citep{barez2025openproblemsmachineunlearning}, or the use of attribution for trustworthy context citations in LM generations~\citep{contextcite, sarti2024quantifying, qi-etal-2024-model}. In this work, signals extracted from model internals are employed to detect errors in models' generated outputs.


\paragraph{Uncertainty Estimation for Language Models}
The estimation of uncertainty in language models has garnered increasing attention~\citep{baan2023uncertaintynaturallanguagegeneration}, particularly in the context of generation tasks for which the set of plausible responses is large~\citep{giulianelli-etal-2023-comes}. Predictive uncertainty is typically decomposed into its \textit{aleatoric} and \textit{epistemic} components, representing respectively the  irreducible variability in the modeled phenomena, and the improvable confidence in model predictions~\citep{aleatoric-epistemic}. Popular methods for uncertainty estimation involve the calibration of predictive probabilities to reflect aleatoric uncertainty \citep{jiang-etal-2020-know,ulmer-etal-2022-exploring,zhao2023calibrating,chen-etal-2023-close}, and conformal sets prediction \citep{zerva-martins-2024-conformalizing,ravfogel-etal-2023-conformal}. In this work, we exploit uncertainty signals from the predictive distribution of MT models and its internal processing for efficiently predicting the resulting generation quality at a fine-grained, token-level scale.

\paragraph{Human Label Variation.} Human label variation is a type of uncertainty that arises from the inherent variability in human judgments \citep{plank-etal-2014-linguistically,plank-2022-problem}, which can be hard to disentangle from actual annotation mistakes~\citep{snow-etal-2008-cheap,weber-genzel-etal-2024-varierr}. The usage of multiple references was recently recommended to ensure a sound evaluation of generative LMs reflecting human-plausible levels of variability \citep{giulianelli-etal-2023-comes}, contrary to common practices employing a single set of ``gold'' labels. In our analysis on QE4PE data containing multiple edits, we adopt a perspectivist approach\footnote{\href{https://pdai.info/}{pdai.info}} to ensure a robust assessment of WQE metrics by accounting for annotators' disagreement~\citep{uma2021learning}.

\section{Details about Models and Datasets}
\label{app:data}

\subsection{MT Models}

\paragraph{mBART-50 1-to-many.} The original multilingual BART (mBART-25) model by \citet{liu-etal-2020-multilingual-denoising} is an encoder-decoder Transformer model pre-trained on monolingual documents in 25 languages with the BART denoising objective for sequence-to-sequence learning~\citep{lewis-etal-2020-bart}. \citet{tang-etal-2021-multilingual} extended mBART-25 by including 25 additional languages during pre-training and performing multilingual translation fine-tuning across 50 languages. In this work, we employ the \textit{one-to-many} version of the model specialized in out-of-English translation that was employed by \citet{sarti-etal-2022-divemt} to produce part of the translations post-edited by DivEMT annotators.\footnote{\href{https://huggingface.co/facebook/mbart-large-50-one-to-many-mmt}{\texttt{facebook/mbart-large-50-one-to-many-mmt}}} The model is a standard Transformer with 12 layers of encoder and 12 layers of decoder, with model dimension of 1024 and 16 attention heads ($\sim$680M parameters).

\paragraph{NLLB 3.3B} (No Language Left Behind) is a collection of multilingual MT models covering up to 202 languages, including low-resource directions \citep{nllb}. The largest NLLB model available is a mixture-of-experts model with 54.4B parameters, which comes with high computational cost. In this work we employ the largest available dense variant of the model ($\sim$3.3B parameters), which was used by \citet{sarti-etal-2025-qe4pe} for collecting the QE4PE post-editing dataset.\footnote{\href{https://huggingface.co/facebook/nllb-200-3.3B}{\texttt{facebook/nllb-200-3.3B}}} The model is an encoder-decoder Transformer with 24 layers for each module, a model dimension of 2048 and 16 attention heads per layer.

\paragraph{Aya23 35B} is a large language model introduced by \citet{aya23} to improve the multilingual capabilities of the original Aya model \citep{ustun-etal-2024-aya} on a selected set of 23 languages. The model was included in the WMT24 evaluation of \citet{kocmi-etal-2024-findings}, resulting in the best translation performances among tested open-source models. The model is a decoder-only Transformer model with 40 layers, a model dimension of 8196 and 64 attention heads per layer.

\subsection{Datasets}

\paragraph{DivEMT} was created by \citep{sarti-etal-2022-divemt} to evaluate the impact of language typology on MT quality, and how that would influence the productivity of human post-editors working with those systems. The dataset includes out-of-English machine translations for Wiki data produced by Google Translate and mBART-50 1-to-many, with edits made by professional translators in six languages. In this work, we evaluate unsupervised metrics on the mBART-50 1-to-many model, converting the human post-edits.

\paragraph{WMT24} employed in this study is taken from the General Machine Translation Shared Task at WMT 2024 \citep{kocmi-etal-2024-findings}. It contains evaluation of several machine translation systems across English$\rightarrow$\{Czech, Hindi, Japanese, Chinese, Russian\} (634 segments) and Czech$\rightarrow$Ukrainian (1954 segments).
The human evaluation was done with the Error Span Annotation protocol (ESA, \citealp{kocmi-etal-2024-error}), which has human annotators highlighting erroneous spans in the translation and marking them as either \textsc{minor} or \textsc{major} errors.
This dataset covers the \textit{news}, \textit{social}, and \textit{speech} (with automatic speech recognition) domains. We adopt the official prompting setup from the WMT24 campaign, using the Aya23 model alongside the provided prompt and 3 in-context translation examples per language to ensure uniformity with previous results.\footnote{\href{https://github.com/wmt-conference/wmt-collect-translations}{\texttt{wmt-conference/wmt-collect-translations}}}

\paragraph{QE4PE}
The QE4PE dataset was created by \citet{sarti-etal-2025-qe4pe} for measuring the effect of word-level error highlights when included in real-world human post-editing workflows. The QE4PE data provides granular behavioral metrics to evaluate the speed and quality of post-editing of 12 annotators for \EnIt and \EnNl across four error span highlighting modalities, including the unsupervised Surprisal MCD\textsubscript{\textsc{var}} method and the supervised \textsc{xcomet-xxl} we also test in this study. Provided that the presence of error span highlights was found to influence the editing choices of human editors, we limit our evaluation to the six human annotators per language that post-edited sentences without any highlights (3 for the \textit{Oracle Post-edit} task to produce initial human-based highlights, and 3 for the \textit{No Highlight} modality in the main task). This prevents us from biasing our evaluation of WQE metrics in favor of the metrics that influenced editing choices.
We use the post-edited versions to synthetically create error spans, which can be used as binary labels to evaluate WQE metrics.

\begin{table*}
    \footscriptsize
    \centering
    \setlength{\tabcolsep}{4.5pt}
    \begin{tabular}{lp{12cm}}
        \toprule[1.5pt] 
        Source\textsubscript{~\textsc{en}} & So why is it that people jump through extra hoops to install Google Maps? \\
        MT\textsubscript{~\textsc{it}} (NLLB) & Quindi perché le persone devono fare un salto in più per installare Google Maps? \\
        \midrule 
        Annotator $t1$ & Quindi perché le persone devono fare un \ncbox{YellowGreen}{salto}{passaggio} in più per installare Google Maps?\\
        Annotator $t2$ & Quindi perché le persone \ncbox{YellowGreen}{devono fare un salto in più}{fanno i salti mortali} per installare Google Maps?\\
        Annotator $t3$ & Quindi perché le persone \ncbox{YellowGreen}{devono fare un salto in più}{effettuano dei passaggi ulteriori e superflui} per installare Google Maps? \\
        Annotator $t4$ & \ncbox{YellowGreen}{Quindi}{Allora} perché le persone \ncbox{YellowGreen}{devono fare}{fanno} un \ncbox{YellowGreen}{salto}{passaggio} in più per installare Google Maps? \\
        Annotator $t5$ & \ncbox{YellowGreen}{Quindi perché le persone devono fare un salto in più}{E allora mi chiedo: perché gli utenti iPhone si affannano tanto} per installare Google Maps? \\
        Annotator $t6$ & Quindi perché le persone \ncbox{YellowGreen}{devono fare un salto in più}{fanno di tutto} per installare Google Maps? \\
        Edit Counts (Fig. \ref{fig:annotators-perf}) & \ncbox{Blue2}{2}{Quindi} \ncbox{Blue1}{1}{perché le persone} \ncbox{Blue5}{5}{devono fare} \ncbox{Blue4}{4}{un} \ncbox{Blue6}{6}{salto} \ncbox{Blue4}{4}{in più} per installare Google Maps? \\
        \midrule
        \textsc{xcomet-xl}       & Quindi perché le persone \ncbox{Red3}{minor}{devono fare} un \ncbox{Red3}{minor}{salto in più} per installare Google Maps? \\
        \textsc{xcomet-xxl}      & \ncbox{Red3}{minor}{Quindi perché} le persone \ncbox{Red5}{major}{devono fare un salto in più} per installare Google Maps? \\
        \textsc{xcomet-xl}\textsubscript{~\textsc{conf}} & \ncbox{Red2}{.41}{Quindi} \ncbox{Red1}{.36}{perché} \ncbox{Red2}{.51}{le} \ncbox{Red2}{.50}{persone} \ncbox{Red4}{.69}{devono} \ncbox{Red4}{.73}{fare} \ncbox{Red2}{.51}{un} \ncbox{Red6}{.81}{salto} \ncbox{Red4}{.74}{in} \ncbox{Red4}{.76}{più} \ncbox{Red1}{.39}{per} \ncbox{Red2}{.47}{install} \ncbox{Red2}{.53}{are} \ncbox{Red1}{.26}{Google} \ncbox{Red1}{.36}{Maps} \ncbox{Red1}{.24}{?} \\
        \textsc{xcomet-xxl}\textsubscript{~\textsc{conf}} & \ncbox{Red2}{.51}{Quindi} \ncbox{Red6}{.83}{perché} \ncbox{Red1}{.20}{le} \ncbox{Red1}{.20}{persone} \ncbox{Red4}{.42}{devono} \ncbox{Red6}{.84}{fare} \ncbox{BrickRed}{.90}{un} \ncbox{BrickRed}{.95}{salto} \ncbox{Red6}{.86}{in} \ncbox{Red6}{.78}{più} \ncbox{White}{.03}{per} \ncbox{White}{.00}{install} \ncbox{White}{.01}{are} \ncbox{White}{.00}{Google} \ncbox{White}{00}{Maps} \ncbox{White}{.00}{?} \\
        Surprisal MCD\textsubscript{~\textsc{var}} & \ncbox{White}{.05}{Quindi} \ncbox{White}{.01}{perché} \ncbox{White}{.04}{le} \ncbox{White}{.00}{persone} \ncbox{Red6}{.41}{devono} \ncbox{Red2}{.09}{fare} \ncbox{White}{.04}{un} \ncbox{BrickRed}{.59}{sal} \ncbox{White}{.00}{to} \ncbox{Red3}{.12}{in} \ncbox{White}{.00}{più} \ncbox{White}{.00}{per} \ncbox{White}{.00}{installare} \ncbox{White}{.00}{Google} \ncbox{White}{.00}{Maps} \ncbox{White}{.00}{?} \\
        \bottomrule[1.5pt] 
    \end{tabular}
    \caption{Annotated example from the \EnIt portion of the QE4PE dataset. \textbf{Top:} Annotator edits with highlighted \colorbox{YellowGreen}{final text} and replaced text on top, with count-based aggregation showing inter-annotator agreement. \textbf{Bottom:} Word-level annotations for best-performing metrics discussed in the study.}
    \label{tab:example-qe4pe-ita}
\end{table*}

\begin{table*}
    \footscriptsize
    \centering
    \setlength{\tabcolsep}{4.5pt}
    \begin{tabular}{lp{12cm}}
        \toprule[1.5pt] 
        Source\textsubscript{~\textsc{en}} & So the challenges in this are already showing themselves. I'm likely going to have a VERY difficult time getting a medical clearance due to the FAA's stance on certain medications. \\
        MT\textsubscript{~\textsc{it}} (Aya23) & Takže problémy s tím se již projevují. Pravděpodobně budu mít PŘESNĚ obtížný čas dostat lékařské potvrzení kvůli postoji FAA k některým lékům. \\
        \midrule 
        Annotator & Takže \ncbox{Blue1}{minor}{problémy} s tím se již projevují. Pravděpodobně budu mít \ncbox{Blue6}{major}{PŘESNĚ obtížný čas} dostat lékařské potvrzení kvůli postoji FAA k některým lékům.\\
        \midrule
        \textsc{xcomet-xl}       & Takže problémy s tím se již projevují. Pravděpodobně budu mít \ncbox{Red3}{minor}{PŘESNĚ obtížný} \ncbox{Red3}{minor}{čas dostat} lékařské \ncbox{Red3}{minor}{potvrzení} kvůli postoji FAA k některým lékům \\
        \textsc{xcomet-xxl}      & \ncbox{Red3}{minor}{Takže problémy s tím se již projevují}. Pravděpodobně budu mít \ncbox{Red5}{major}{PŘESNĚ obtížný čas dostat} lékařské potvrzení kvůli postoji FAA k některým lékům. \\
        \textsc{xcomet-xl}\textsubscript{~\textsc{conf}} & \ncbox{Red3}{0.23}{Takže}
        \ncbox{Red4}{0.28}{problémy} \ncbox{Red3}{0.26}{s} \ncbox{Red4}{0.28}{tím} \ncbox{Red2}{0.17}{se} \ncbox{Red2}{0.19}{již} \ncbox{Red3}{0.31}{projevují} \ncbox{Red2}{0.17}{.} \ncbox{Red3}{0.23}{Pravděpodobně} \ncbox{Red5}{0.40}{budu} \ncbox{Red5}{0.48}{mít} \ncbox{BrickRed}{0.79}{PŘESNĚ} \ncbox{Red6}{0.65}{obtížný} \ncbox{Red6}{0.76}{čas} \ncbox{Red6}{0.64}{dostat} \ncbox{Red5}{0.50}{lékařské} \ncbox{Red5}{0.51}{potvrzení} \ncbox{Red2}{0.19}{kvůli} \ncbox{Red4}{0.34}{postoji} \ncbox{Red3}{0.27}{FAA} \ncbox{Red3}{0.20}{k} \ncbox{Red3}{0.20}{některým} \ncbox{Red3}{0.21}{lékům} \ncbox{Red2}{0.17}{.} \\
        \textsc{xcomet-xxl}\textsubscript{~\textsc{conf}} & \ncbox{Red5}{0.25}{Takže} \ncbox{Red4}{0.24}{problémy} \ncbox{Red4}{0.26}{s} \ncbox{Red5}{0.31}{tím} \ncbox{Red5}{0.29}{se} \ncbox{Red4}{0.23}{již} \ncbox{Red5}{0.26}{projevují} \ncbox{White}{0.01}{.} \ncbox{White}{0.01}{Pravděpodobně} \ncbox{White}{0.03}{budu} \ncbox{Red6}{0.37}{PŘESNĚ} \ncbox{Red5}{0.30}{obtížný} \ncbox{Red5}{0.32}{čas} \ncbox{Red4}{0.24}{dostat} \ncbox{Red3}{0.10}{lékařské} \ncbox{Red4}{0.13}{potvrzení} \ncbox{White}{0.01}{kvůli} \ncbox{White}{0.00}{postoji} \ncbox{White}{0.00}{FAA} \ncbox{White}{0.00}{k} \ncbox{White}{0.00}{některým} \ncbox{White}{0.00}{lékům} \ncbox{White}{0.00}{.}
 \\
        Out. Entropy & \ncbox{Red4}{0.88}{Takže} \ncbox{Red6}{1.93}{problémy} \ncbox{Red6}{1.88}{s} \ncbox{Red3}{0.84}{tím} \ncbox{Red5}{1.66}{se} \ncbox{Red3}{1.13}{již} \ncbox{Red4}{0.89}{projevují} \ncbox{White}{0.11}{.} \ncbox{Red2}{0.44}{Pravděpodobně} \ncbox{Red1}{0.22}{budu} \ncbox{White}{0.09}{mít} \ncbox{BrickRed}{2.09}{PŘESNĚ} \ncbox{BrickRed}{3.70}{obtížný} \ncbox{White}{0.09}{čas} \ncbox{Red6}{1.40}{dostat} \ncbox{Red5}{1.02}{lékařské} \ncbox{Red3}{0.64}{potvrzení} \ncbox{Red2}{0.69}{kvůli} \ncbox{Red2}{0.24}{postoji} \ncbox{Red3}{0.80}{FAA} \ncbox{Red5}{1.01}{k} \ncbox{Red3}{0.55}{některým} \ncbox{Red1}{0.18}{lékům} \ncbox{White}{0.11}{.}
 \\
        \bottomrule[1.5pt] 
    \end{tabular}
    \caption{Annotated example from the \EnCs portion of the WMT24 dataset. \textbf{Top:} Annotator edits with highlighted Error Span Annotation of \colorbox{Blue1}{minor} and \colorbox{Blue6}{major} errors. \textbf{Bottom:} Word-level annotations for best-performing metrics discussed in the study.}
    \label{tab:example-wmt24-encs}
\end{table*}

\section{Details about Tested Metrics}

\paragraph{Monte Carlo Dropout (MCD)} is a technique introduced by \citet{mcdropout} for estimating model uncertainty at inference time. MCD uses the dropout mechanism in neural networks~\citep{dropout}, commonly employed for regularization during training, at inference time to produce a set of noisy predictions from a unique model, approximating Bayesian inference. For a given input $x$, $T$ forward passes are performed through the network. In each pass $t \in T$, a different random dropout mask $\Theta_t$ is applied, resulting in a slightly different output probabilities $p(x \mid \Theta_t)$. The set of $T$ predictions $\{p(x \mid \Theta_1), \dots, p(x \mid \Theta_T)\}$ can be seen as samples from an approximate posterior distribution. In this work, we employ the mean of the negative log probabilities as a robust estimate of surprisal:
$$\text{Surprisal MCD\textsubscript{avg}} = \hat y_\text{\textsubscript{MCD}} = \frac{1}{T} \sum_{t=1}^{T} - \log p(x | \Theta_t)$$
Moreover, we estimate predictive uncertainty by calculating the variance of predictive probabilities under the same setup:
$$\text{Surprisal MCD\textsubscript{var}} = \frac{1}{T} \sum_{t=1}^{T} \big(- \log p(x | \Theta_t) - \hat y_\text{\textsubscript{MCD}} \big)$$

\paragraph{Vocabulary Projections.} The Logit Lens \citep{logitlens} is an interpretability technique used to understand the internal workings of Transformer models, particularly how their predictions evolve layer by layer. Activations $h_l$ produced by the model layer $l$ are projected to vocabulary space using the model unembedding matrix, $W_U$, normally used to produce output logits. For the NLLB and mBART-50 models we also apply a final layer normalization before the projection, following the model architecture, while for the Aya model we scale logits by $0.0625$ (the default \texttt{logit\_scale} defined in the model configuration). Following the residual stream view of the Transformer model~\citep{elhage2021mathematical}, the resulting logits provide a view into the model predictive confidence at that specific depth of processing.

\paragraph{Context mixing.} Several works studied the mixing of contextual information across language model layers to attribute model predictions to specific input properties~(\citealp{ferrando-etal-2022-measuring,mohebbi-etal-2023-quantifying,ferrando-etal-2023-explaining} \textit{inter alia}). In this work we employ simple estimates of context relevance using attention weights produced during the Transformer attention operation. More specifically, for every attention head at every layer of the decoder module, we extract a score for every token in the preceding context, employing cross-attention weights to account for source-side context in encoder-decoder models.

\paragraph{\textsc{xcomet}} is a suite of MT evaluation metrics introduced by \citet{guerreiro-etal-2024-xcomet} extending the popular \textsc{comet} metric \citep{rei-etal-2020-comet} to combine combines sentence-level and word-level error
span prediction for improved explainability of results. \textsc{xcomet} metrics come in a 3B (XL) and 11B (XXL) size and support both reference-based and reference-less usage, hence enabling usage for quality estimation purposes. Concretely, \textsc{xcomet} models are Transformer encoders fine-tuned from pre-trained XLMR encoders~\citep{goyal-etal-2021-larger} using a mix of sentence-level Direct Assessment scores and word-level MQM error spans. In this work, we focus on their word-level error span prediction capabilities in a quality estimation setup, where the model classifies every input token according to MQM severity levels $\{$\textsc{ok, minor, major, critical}$\}$ with a learned linear layer.\footnote{The default \textsc{xcomet} metric was used with the \texttt{unbabel-comet} library (\texttt{v2.2.6}).}

\paragraph{Token-level Evaluation.} Error spans used as labels in our evaluation are defined at the character level, while metric scores depend on the tokenization employed by either the MT model (for unsupervised metrics) or \textsc{xcomet} (for supervised metrics). To allow for comparison, we label tokens as being part of an error span if at least one character contained in it was marked as an error or edited by an annotator. \Cref{tab:example-qe4pe-ita,tab:example-wmt24-encs} provide examples of various segmentations for the same MT output.

\paragraph{Constraining generation} Evaluating metrics at the word level can be challenging due to the need for perfect uniformity between model generations and annotated spans. For this reason we extract unsupervised metrics during generation while force-decoding the annotated outputs from the MT model to ensure perfect adherence with annotated error spans. In general, such an approach could introduce a problematic confounder in the evaluation, since observed results could be the product of constraining a model towards an unnatural generation, rather than reflecting underlying phenomena. However, in this study, we carefully ensure that the generation setup matches exactly the one of previous works where the annotated translations were produced, using the same MT model and the same inputs.\footnote{Generation parameters are not relevant in this setting, provided that they only alter the selection of the following output token, which we do via force-decoding.} Hence, the constraining process is a simple insurance of conformity in light of potential discrepancies introduced by different decoding strategies, and does not affect the soundness of our method.

\begin{table*}
    \small
    \centering
    \begin{tabular}{l|cc|cc|cc|cc|cc|cc||cc}
        \toprule[1.5pt]
        \multirow{2}{*}{\textbf{Method}} & \multicolumn{2}{c}{\textbf{QE4PE}$_{\mathbf{t1}}$} & \multicolumn{2}{c}{\textbf{QE4PE}$_{\mathbf{t2}}$} & \multicolumn{2}{c}{\textbf{QE4PE}$_{\mathbf{t3}}$} & \multicolumn{2}{c}{\textbf{QE4PE}$_{\mathbf{t4}}$} & \multicolumn{2}{c}{\textbf{QE4PE}$_{\mathbf{t5}}$} & \multicolumn{2}{c}{\textbf{QE4PE}$_{\mathbf{t6}}$} & \multicolumn{2}{c}{\textbf{QE4PE}$_{\mathbf{avg}}$}\\

        \cmidrule(lr){2-3}
        \cmidrule(lr){4-5}
        \cmidrule(lr){6-7}
        \cmidrule(lr){8-9}
        \cmidrule(lr){10-11}
        \cmidrule(lr){12-13}
        \cmidrule(lr){14-15}

                                  & \b{AP}  & \b{F1$^{*}$}  & \b{AP}  & \b{F1$^{*}$}  & \b{AP}  & \b{F1$^{*}$}  & \b{AP}  & \b{F1$^{*}$}  & \b{AP}  & \b{F1$^{*}$}  & \b{AP}  & \b{F1$^{*}$}  & \b{AP}  & \b{F1$^{*}$} \\

        \midrule
        Random Baseline           & .08     & .14     & .15     & .26     & .06     & .12     & .11     & .19     & .22     & .36     & .18     & .30     & .13     & .23     \\
        \midrule
        Surprisal                 & .11     & .20     & .21     & .31     & .11     & .17     & .16     & .25     & .30     & .40     & .25     & .35     & .19     & .28     \\
        Out. Entropy              & .12     & .18     & .22     & .30     & .10     & .16     & .17     & .24     & .30     & .39     & .26     & .34     & .19     & .27     \\
        Surprisal MCD\avg         & .12     & .20     & .22     & .32     & .11     & .17     & .16     & .26     & .30     & \b{\u{.41}} & .26     & \u{.36} & .19     & .29     \\
        Surprisal MCD\var         & \u{.13} & \u{.21} & \u{.26} & \u{.33} & \u{.12} & \u{.20} & \u{.19} & \u{.27} & \u{.31} & .40     & \u{.29} & \u{.36} & \u{.22} & \u{.30} \\
        LL Surprisal\best         & .11     & .19     & .21     & .32     & .11     & .16     & .16     & .25     & .29     & .40     & .26     & .35     & .19     & .28     \\
        LL KL-Div\best            & .09     & .16     & .19     & .28     & .08     & .14     & .13     & .21     & .25     & .37     & .22     & .31     & .16     & .25     \\
        LL Pred. Depth            & .09     & .16     & .18     & .28     & .07     & .13     & .14     & .21     & .25     & .37     & .21     & .31     & .16     & .24     \\
        Attn. Entropy\avg         & .11     & .16     & .17     & .27     & \u{.12} & .17     & .11     & .19     & .23     & .36     & .19     & .31     & .15     & .24     \\
        Attn. Entropy\maxim       & .09     & .14     & .15     & .26     & .10     & .18     & .09     & .19     & .20     & .36     & .16     & .30     & .13     & .24     \\
        \textsc{Blood}\best       & .08     & .14     & .16     & .26     & .06     & .12     & .11     & .19     & .23     & .36     & .18     & .30     & .14     & .23     \\
        \midrule
        \textsc{xcomet-xl}        & .11     & .24     & .22     & .35     & .10     & .20     & .16     & .30     & .27     & .35     & .23     & .34     & .18     & .30     \\
        \textsc{xcomet-xl}\confw  & \b{.20} & .25     & .30     & \b{.36} & .14     & .21     & .25     & .31     & \b{.37} & .40     & .31     & .36     & .26     & .32     \\
        \textsc{xcomet-xxl}       & .13     & \b{.27} & .22     & .32     & .10     & \b{.24} & .17     & .31     & .28     & .32     & .23     & .31     & .19     & .30     \\
        \textsc{xcomet-xxl}\confw & .19     & \b{.27} & \b{.31} & \b{.36} & \b{.17} & \b{.24} & \b{.26} & \b{.32} & \b{.37} & \b{.41} & \b{.33} & \b{.39} & \b{.27} & \b{.33} \\
        \midrule
        Human Editors\minim       & .17     & .33     & .26     & .38     & .10     & .21     & .16     & .26     & .25     & .36     & .23     & .30     & .19     & .31     \\
        Human Editors\avg         & .20     & .38     & .29     & .43     & .14     & .30     & .22     & .39     & .32     & .38     & .30     & .40     & .25     & .39     \\
        Human Editors\maxim       & .24     & .43     & .31     & .47     & .20     & .41     & .24     & .43     & .37     & .50     & .33     & .50     & .28     & .46     \\ 
        \bottomrule[1.5pt]
    \end{tabular}
    \caption{WQE metrics' performance for predicting error spans from the six edit sets over NLLB 3.3B translations in the \EnIt QE4PE dataset \citep{sarti-etal-2025-qe4pe}. \u{Best unsupervised} and \b{overall best} metric results are highlighted.}
    \label{tab:qe4pe-ita-all-results}
\end{table*}

\begin{table*}
    \small
    \centering
    \begin{tabular}{l|cc|cc|cc|cc|cc|cc||cc}
        \toprule[1.5pt]
        \multirow{2}{*}{\textbf{Method}} & \multicolumn{2}{c}{\textbf{QE4PE}$_{\mathbf{t1}}$} & \multicolumn{2}{c}{\textbf{QE4PE}$_{\mathbf{t2}}$} & \multicolumn{2}{c}{\textbf{QE4PE}$_{\mathbf{t3}}$} & \multicolumn{2}{c}{\textbf{QE4PE}$_{\mathbf{t4}}$} & \multicolumn{2}{c}{\textbf{QE4PE}$_{\mathbf{t5}}$} & \multicolumn{2}{c}{\textbf{QE4PE}$_{\mathbf{t6}}$} & \multicolumn{2}{c}{\textbf{QE4PE}$_{\mathbf{avg}}$}\\

        \cmidrule(lr){2-3}
        \cmidrule(lr){4-5}
        \cmidrule(lr){6-7}
        \cmidrule(lr){8-9}
        \cmidrule(lr){10-11}
        \cmidrule(lr){12-13}
        \cmidrule(lr){14-15}

                                     & \b{AP}  & \b{F1$^{*}$}  & \b{AP}  & \b{F1$^{*}$}  & \b{AP}  & \b{F1$^{*}$}  & \b{AP}  & \b{F1$^{*}$}  & \b{AP}  & \b{F1$^{*}$}  & \b{AP}  & \b{F1$^{*}$}  & \b{AP}  & \b{F1$^{*}$} \\

        \midrule
        Random Baseline              & .07     & .14     & .34     & .51     & .22     & .36     & .19     & .32     & .13     & .24     & .22     & .36     & .20     & .32     \\
        \midrule
        Surprisal                    & .12     & .19     & .41     & .51     & .30     & .39     & .29     & .37     & .21     & .30     & .31     & .41     & .27     & .36     \\
        Out. Entropy                 & .11     & .18     & .41     & .51     & .31     & .37     & .29     & .36     & .20     & .27     & .31     & .39     & .27     & .35     \\
        Surprisal\mcdavg             & .12     & .19     & .42     & .52     & .31     & .40     & .30     & \u{.40}     & .21     & .30     & .31     & \u{.42} & .28     & .37     \\
        Surprisal\mcdvar             & \u{.13} & \u{.21} & \u{.45} & \b{\u{.53}} & \u{.36} & \u{.41} & \u{.34} & \u{.40} & \u{24} & \u{.32} & \u{.36} & \u{.42} & \u{.31} & \u{.38} \\
        LL Surprisal\best            & .12     & .19     & .42     & \b{\u{.53}} & .30     & .40     & .29     & .38     & .21     & .30     & .31     & .41     & .27     & .37     \\
        LL KL-Div\best               & .09     & .15     & .39     & .52     & .28     & .37     & .25     & .34     & .17     & .26     & .29     & .38     & .25     & .34     \\
        LL Pred. Depth               & .09     & .16     & .37     & .52     & .26     & .37     & .24     & .33     & .17     & .25     & .27     & .38     & .23     & .33     \\
        Attn. Entropy\avg            & .09     & .15     & .37     & .51     & .22     & .36     & .20     & .32     & .13     & .24     & .23     & .37     & .21     & .32     \\
        Attn. Entropy\maxim          & .09     & .15     & .35     & .51     & .22     & .36     & .18     & .32     & .12     & .24     & .21     & .37     & .19     & .32     \\
        \textsc{Blood}\best          & .07     & .13     & .35     & .51     & .22     & .36     & .19     & .32     & .14     & .24     & .23     & .36     & .20     & .32     \\
        \midrule
        \textsc{xcomet-xl}           & .13     & .27     & .39     & .39     & .31     & .44     & .28     & .32     & .20     & .35     & .31     & .44     & .27     & .38     \\
        \textsc{xcomet-xl}\confw     & \b{.24} & \b{.31} & .47     & \b{.53} & \b{.43} & \b{.45} & \b{.40} & \b{.43} & .29     & \b{.36} & \b{.43} & \b{.46} & \b{.38} & \b{.42} \\
        \textsc{xcomet-xxl}          & .13     & .28     & .39     & .29     & .30     & .35     & .26     & .35     & .19     & .31     & .30     & .35     & .26     & .32     \\
        \textsc{xcomet-xxl}\confw    & \b{.24} & .30     & \b{.48} & \b{.53} & \b{.43} & \b{.45} & \b{.40} & .42     & \b{.31} & .35     & \b{.43} & .45     & \b{.38} & \b{.42} \\
        \midrule
        Human Editors\minim          & .16     & .29     & .43     & .51     & .34     & .45     & .33     & .47     & .26     & .42     & .36     & .46     & .32     & .43     \\
        Human Editors\avg            & .17     & .33     & .44     & .51     & .34     & .45     & .33     & .47     & .26     & .42     & .36     & .46     & .32     & .43     \\
        Human Editors\maxim          & .19     & .36     & .46     & .51     & .36     & .51     & .37     & .53     & .32     & .51     & .40     & .53     & .35     & .49     \\ 
        \bottomrule[1.5pt]
    \end{tabular}
    \caption{WQE metrics' performance for predicting error spans from the six edit sets over NLLB 3.3B translations in the \EnNl QE4PE dataset \citep{sarti-etal-2025-qe4pe}. \u{Best unsupervised} and \b{overall best} metric results are highlighted.}
    \label{tab:qe4pe-nld-all-results}
\end{table*}

\begin{table*}
\footnotesize
\centering
\begin{tabular}{l|cc|cc|cc|cc|cc|cc||cc}
\toprule[1.5pt]
\multicolumn{1}{c}{\multirow{2}{*}{\textbf{Method}}} & \multicolumn{2}{c}{\textbf{Italian}} & \multicolumn{2}{c}{\textbf{Dutch}} & \multicolumn{2}{c}{\textbf{Arabic}} & \multicolumn{2}{c}{\textbf{Turkish}} & \multicolumn{2}{c}{\textbf{Vietnamese}} & \multicolumn{2}{c}{\textbf{Ukrainian}} & \multicolumn{2}{c}{\textbf{Average}}\\
\cmidrule(lr){2-3}
\cmidrule(lr){4-5}
\cmidrule(lr){6-7}
\cmidrule(lr){8-9}
\cmidrule(lr){10-11}
\cmidrule(lr){12-13}
\cmidrule(lr){14-15}

                                     & \b{AP}  & \b{F1$^{*}$}  & \b{AP}  & \b{F1$^{*}$}  & \b{AP}  & \b{F1$^{*}$}  & \b{AP}  & \b{F1$^{*}$}  & \b{AP}  & \b{F1$^{*}$}  & \b{AP}  & \b{F1$^{*}$}  & \b{AP}  & \b{F1$^{*}$} \\
\midrule
Random Baseline           &    .25  & .40     &    .28  & .43     &    .33  & .49     &    .34  & .50     &    .35  & .52     &    .48  & .65     &    .34  & .50     \\
\midrule
Surprisal                 &    .34  & .45 &    .36  & .46 &    .42  & .51     &    .43  & .54 & .46 & \u{.55} &    .55  & .65     &    .43  & .53 \\
Out. Entropy              & .37 & .43     & .39 & .45     &    .45  & .50     & \u{.49} & .52     & \u{.48} & .54     & .58 & .65     & .46 & .51     \\
Surprisal\mcdavg          &    .34  & .45 &    .37  & \u{.47} &    .43  & .52     &    .44  & .54 & .46 & \u{.55} &    .56  & .65     &    .43  & .53 \\
Surprisal\mcdvar          & \u{.39} & \u{.46} & \u{.41} & \u{.47} & \u{.47} & \u{.53} & \u{.49} & \u{.55} & \u{.48} & \u{.55} & \u{.61} & \b{\u{.67}} & \u{.48} & \u{.54} \\
LL Surprisal\best         &    .33  & .44     &    .36  & .45     &    .41  & .51     &    .44  & .54 &    .44  & \u{.55} &    .55  & .66 &    .42  & .53 \\
LL KL-Div\best            &    .34  & .42     &    .37  & .45     &    .41  & .51     &    .44  & .52     &    .44  & .52     &    .56  & .65     &    .43  & .51     \\
LL Pred. Depth            &    .30  & .42     &    .32  & .44     &    .39  & .50     &    .40  & .52     &    .39  & .53     &    .54  & .66 &    .39  & .51     \\
Attn. Entropy\avg         &    .28  & .41     &    .30  & .43     &    .35  & .49     &    .37  & .51     &    .40  & .52     &    .50  & .65     &    .37  & .50     \\
Attn. Entropy\maxim       &    .25  & .41     &    .26  & .43     &    .34  & .49     &    .34  & .50     &    .35  & .52     &    .47  & .65     &    .34  & .50     \\
\textsc{Blood}\best       &    .26  & .40     &    .28  & .43     &    .35  & .52 &    .35  & .50     &    .36  & .52     &    .49  & .65     &    .35  & .51     \\
\midrule
\textsc{xcomet-xl}        &    .34  & .39     &    .37  & .44     &    .41  & .47     &    .44  & .50     &    .42  & .44     &    .56  & .44     &    .42  & .45     \\
\textsc{xcomet-xl}\confw  &    .46  & .47     &    .49  & \b{.50} &    .51  & .53     & \b{.58} & \b{.56} &    .53  & .55     &    .68  & \b{.67} &    .54  & \b{.55} \\
\textsc{xcomet-xxl}       &    .34  & .36     &    .35  & .35     &    .43  & .47     &    .45  & .48     &    .43  & .42     &    .57  & .41     &    .43  & .42     \\
\textsc{xcomet-xxl}\confw & \b{.48} & \b{.49} & \b{.50} & \b{.50} & \b{.55} & \b{.54} & \b{.58} & \b{.56} & \b{.56} & \b{.57} & \b{.70} & \b{.67} & \b{.56} & \b{.55} \\

\bottomrule[1.5pt]
\end{tabular}
\caption{WQE metrics' performance for predicting error spans from multiple edit sets (one per language) over mBART-50 translations across the six topologically diverse target languages of \textsc{DivEMT} \citep{sarti-etal-2022-divemt}.}
\label{tab:divemt-all-results}
\end{table*}

\begin{table*}
\footnotesize
\centering
\begin{tabular}{l|cc|cc|cc|cc|cc|cc||cc}
\toprule[1.5pt]
\multicolumn{1}{c}{\multirow{2}{*}{\textbf{Method}}} & \multicolumn{2}{c}{\textbf{En$\rightarrow$ Ja}} & \multicolumn{2}{c}{\textbf{En$\rightarrow$ Zh}} & \multicolumn{2}{c}{\textbf{En$\rightarrow$ Hi}} & \multicolumn{2}{c}{\textbf{Cs$\rightarrow$ Uk}} & \multicolumn{2}{c}{\textbf{En$\rightarrow$ Cs}} & \multicolumn{2}{c}{\textbf{En$\rightarrow$ Ru}} & \multicolumn{2}{c}{\textbf{Average}}\\
\cmidrule(lr){2-3}
\cmidrule(lr){4-5}
\cmidrule(lr){6-7}
\cmidrule(lr){8-9}
\cmidrule(lr){10-11}
\cmidrule(lr){12-13}
\cmidrule(lr){14-15}

                                     & \b{AP}  & \b{F1$^{*}$}  & \b{AP}  & \b{F1$^{*}$}  & \b{AP}  & \b{F1$^{*}$}  & \b{AP}  & \b{F1$^{*}$}  & \b{AP}  & \b{F1$^{*}$}  & \b{AP}  & \b{F1$^{*}$}  & \b{AP}  & \b{F1$^{*}$} \\
\midrule
Random Baseline              & .02     & .03     & .03     & .07     & .03     & .07     & .05     & .09     & .06     & .11     & .08     & .16     & .05     & .09     \\
\midrule
Surprisal                    & \u{.03} & .07     & .05     & .09     & .05     & .09     & .14     & .20     & .10     & .16     & .13     & .19     & .08     & .13     \\
Out. Entropy                 & \u{.03} & \u{.08} & \u{.06} & \u{.11} & \u{.06} & \u{.10} & \b{\u{.20}} & \u{.27} & \u{.12} & \u{.18} & \u{.14} & \u{.20} & \u{.10} & \u{.16} \\
LL Surprisal\best            & \u{.03} & .07     & .05     & .09     & .05     & .09     & .14     & .20     & .10     & .16     & .13     & .19     & .08     & .13     \\
LL KL-Div\best               & .02     & .05     & .04     & .07     & .04     & .08     & .10     & .17     & .09     & .15     & .12     & .19     & .07     & .12     \\
LL Pred. Depth               & .02     & .05     & .04     & .08     & .04     & .09     & .09     & .18     & .08     & .14     & .11     & .18     & .06     & .12     \\
Attn. Entropy\avg            & .02     & .03     & .03     & .07     & .03     & .07     & .03     & .09     & .05     & .11     & .07     & .16     & .04     & .09     \\
Attn. Entropy\maxim          & .01     & .03     & .03     & .07     & .03     & .07     & .03     & .09     & .05     & .11     & .08     & .16     & .04     & .09     \\
\midrule
\textsc{xcomet-xl}           & .04     & .09     & .05     & .11     & .06     & .12     & .13     & .28     & .11     & .24     & .16     & .32     & .09     & .19     \\
\textsc{xcomet-xl}\confw     & \b{.08} & .14     & \b{.10} & .16     & \b{.10} & \b{.19} & .18     & \b{.30} & .19     & .29     & .24     & .32     & .15     & .23     \\
\textsc{xcomet-xxl}          & .04     & .11     & .06     & .13     & .05     & .11     & .13     & .28     & .11     & .24     & .16     & \b{.33} & .09     & .20     \\
\textsc{xcomet-xxl}\confw    & .07     & \b{.15} & .09     & \b{.19} & .09     & .17     & .19     & .29     & \b{.22} & \b{.30} & \b{.28} & \b{.33} & \b{.16} & \b{.24} \\
\bottomrule[1.5pt]
\end{tabular}
\caption{WQE metrics' performance for predicting error spans from the ESA annotations (one set per language) over Aya23-35B outputs for the WMT24 dataset \citep{kocmi-etal-2024-findings}.}
\label{tab:wmt24esa-all-results}
\end{table*}

\begin{figure*}
    \centering
    \includegraphics[width=0.75\textwidth]{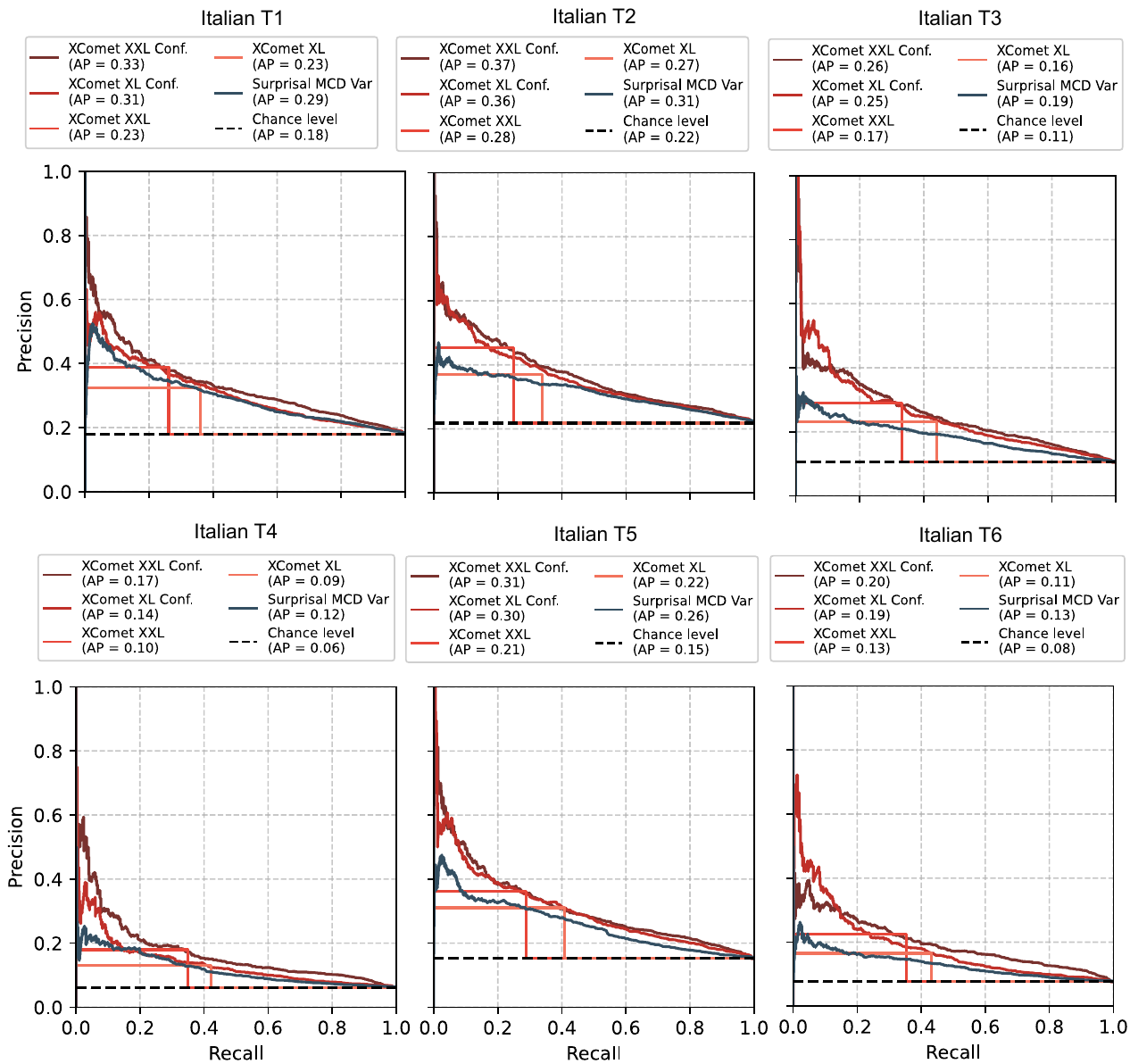}
    \caption{Precision-recall curves for \textsc{xcomet} metrics and Surprisal MCD\textsubscript{\textsc{var}} for all annotators of QE4PE \EnIt.}
    \label{fig:qe4pe-ita-pr-curves}
\end{figure*}

\begin{figure*}
    \centering
    \includegraphics[width=0.75\textwidth]{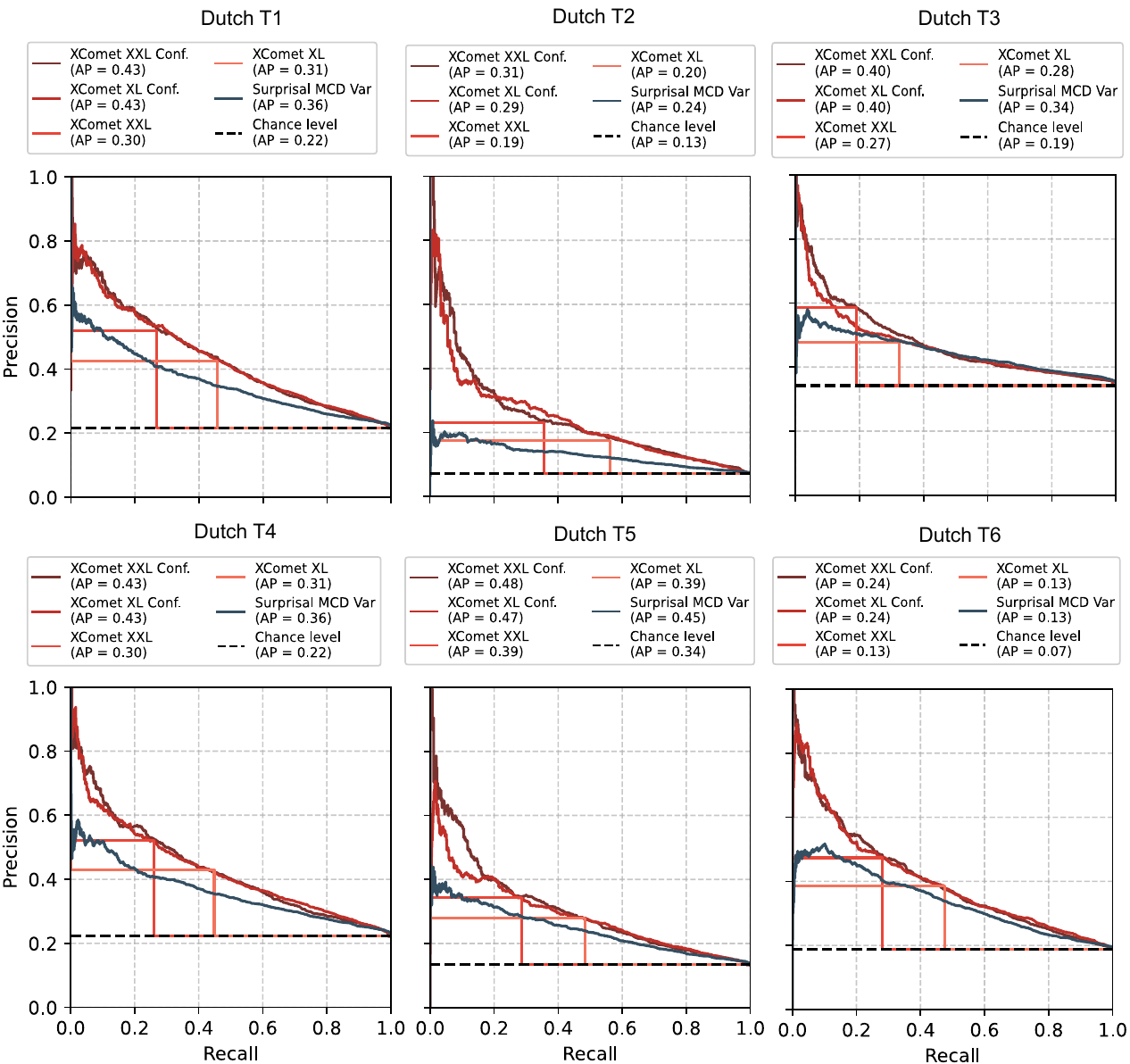}
    \caption{Precision-recall curves for \textsc{xcomet} metrics and Surprisal MCD\textsubscript{\textsc{var}} for all annotators of QE4PE \EnNl.}
    \label{fig:qe4pe-nld-pr-curves}
\end{figure*}

\begin{figure*}
    \centering
    \includegraphics[width=0.75\textwidth]{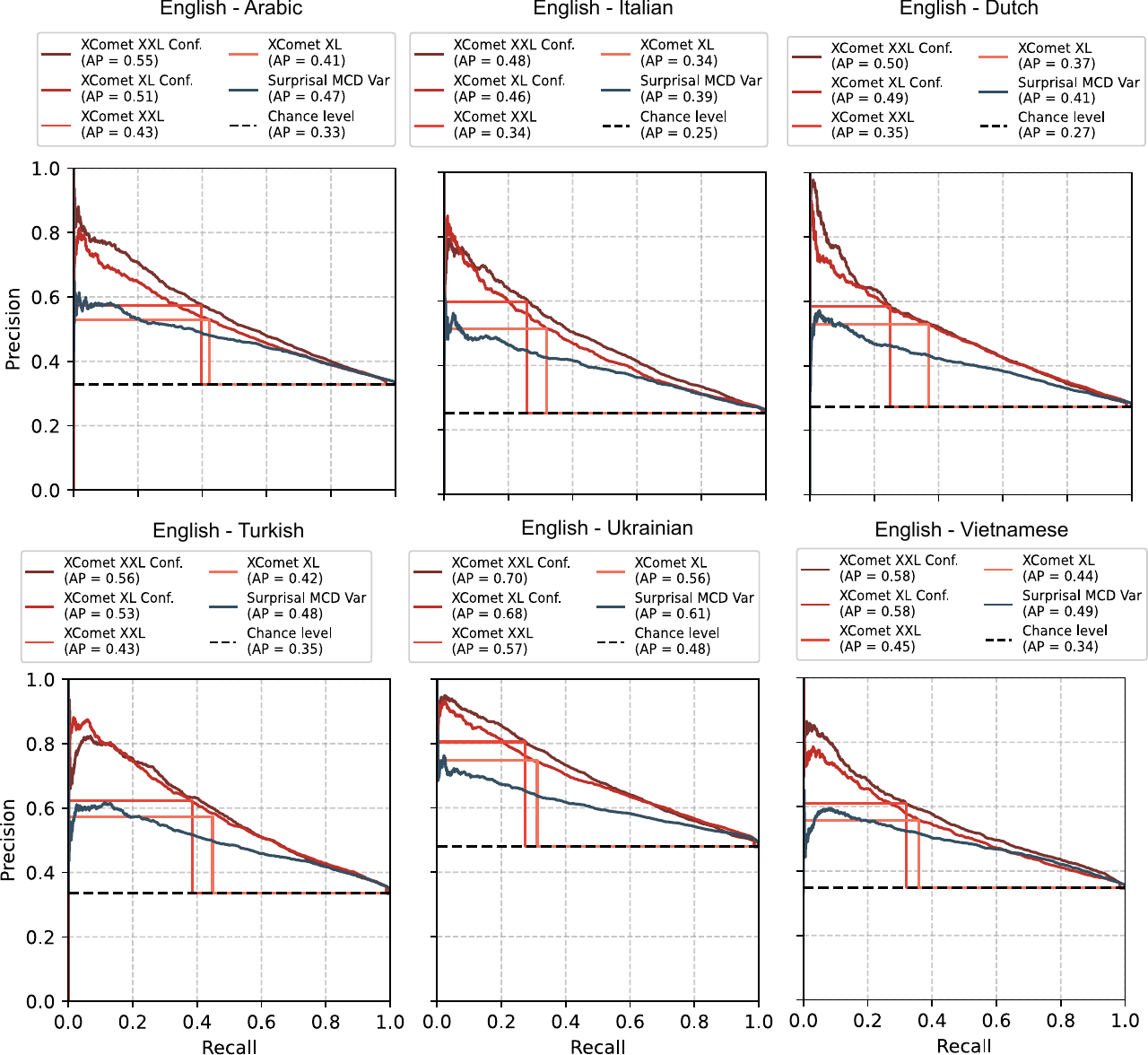}
    \caption{Precision-recall curves for \textsc{xcomet} metrics and Surprisal MCD\textsubscript{\textsc{var}} on all \textsc{DivEMT} languages.}
    \label{fig:divemt-pr-curves}
\end{figure*}

\begin{figure*}
    \centering
    \includegraphics[width=0.75\textwidth]{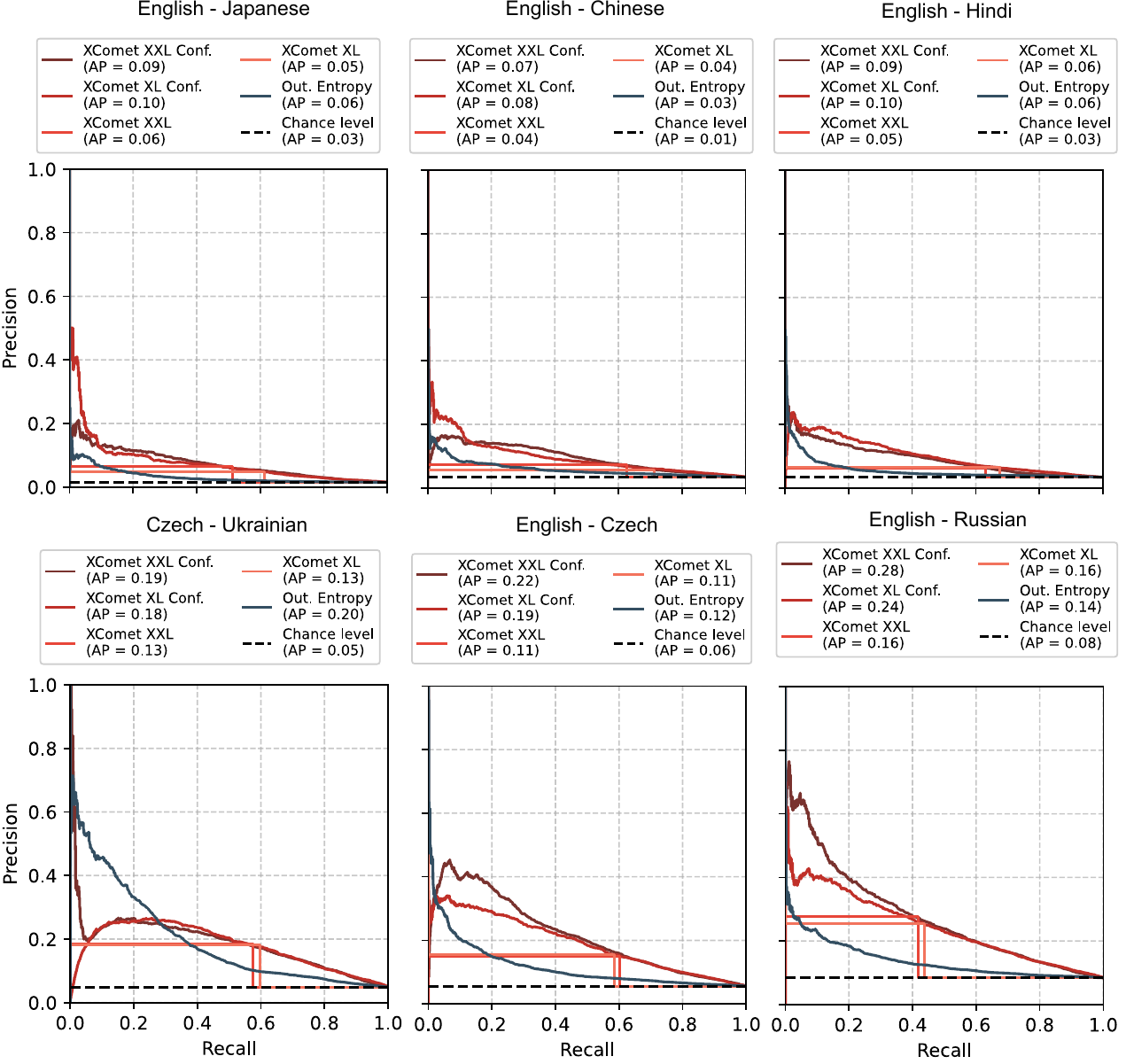}
    \caption{Precision-recall curves for \textsc{xcomet} metrics and Out. Entropy on all WMT24 languages.}
    \label{fig:wmt24esa-pr-curves}
\end{figure*}

\end{document}

%% file: macros.tex
\definecolor{TodoColor}{rgb}{1,0.7,0.6}
\definecolor{TodoColor2}{rgb}{0.7,0.7,0.9}
\definecolor{TodoColor3}{rgb}{0.5,0.8,0.5}

\newcommand{\best}{{\tiny~$_{\textsc{best}}$~}}
\newcommand{\mcdavg}{{\tiny~$_{\textsc{mcd avg}}$~}}
\newcommand{\mcdvar}{{\tiny~$_{\textsc{mcd var}}$~}}
\newcommand{\minim}{{\tiny~$_{\textsc{min}}$~}}
\newcommand{\maxim}{{\tiny~$_{\textsc{max}}$~}}
\newcommand{\var}{{\tiny~$_{\textsc{var}}$~}}
\newcommand{\avg}{{\tiny~$_{\textsc{avg}}$~}}

\newcommand{\confw}{{\tiny~$_{\textsc{conf}}$~}}

\renewcommand{\u}[1]{\underline{#1}}
\renewcommand{\b}[1]{\textbf{#1}}

\newcommand{\EnNl}{\textsc{en}$\rightarrow$\textsc{nl}\xspace}
\newcommand{\EnIt}{\textsc{en}$\rightarrow$\textsc{it}\xspace}

\newcommand{\EnCs}{\textsc{en}$\rightarrow$\textsc{cs}\xspace}

\newcommand{\CsUk}{\textsc{cs}$\rightarrow$\textsc{uk}\xspace}

\let\svthefootnote\thefootnote
\newcommand\blankfootnote[1]{%
  \let\thefootnote\relax\footnotetext{#1}%
  \let\thefootnote\svthefootnote%
}

\definecolor{FindingColor}{HTML}{3080c0}
\newcommand{\finding}[1]{\textcolor{FindingColor}{\bf #1}}

\newcommand{\ncbox}[3]{$\overset{\text{#2}}{\colorbox{#1}{\text{#3}}}$}

\definecolor{Blue1}{RGB}{204, 229, 255}  
\definecolor{Blue2}{RGB}{179, 217, 255}  
\definecolor{Blue3}{RGB}{153, 204, 255}  
\definecolor{Blue4}{RGB}{128, 191, 255}  
\definecolor{Blue5}{RGB}{102, 178, 255}  
\definecolor{Blue6}{RGB}{77, 153, 230}

\definecolor{Red1}{RGB}{255, 229, 229}  
\definecolor{Red2}{RGB}{255, 204, 204}  
\definecolor{Red3}{RGB}{255, 179, 179}  
\definecolor{Red4}{RGB}{255, 153, 153}  
\definecolor{Red5}{RGB}{244, 143, 143}  
\definecolor{Red6}{RGB}{230, 120, 120}

\makeatletter
\newcommand{\footscriptsize}{\fontsize{7.5}{9}\selectfont} 
\patchcmd{\@footnotetext}
  {\footnotesize}
  {\mycustomfootnotesize}
  {}{}
\makeatother

\newcommand{\hlc}[2][yellow]{{%
    \colorlet{foo}{#1}%
    \sethlcolor{foo}\hl{#2}}%
}